\documentclass[sigconf]{acmart}
\AtBeginDocument{%
  }

\author{Haoran Liu}
\authornote{Work done during Haoran's internship at Amazon.}
\affiliation{
    \institution{Texas A\&M University}
    \country{} 
}
\email{liuhr99@tamu.edu}

\author{
    Amir Tahmasbi
}
\affiliation{
    \institution{Amazon Inc.}
    \country{} 
}
\email{
    atahmasb@amazon.com
}

\author{
   Ehtesham Sam Haque
}
\affiliation{
    \institution{Amazon Inc.}
    \country{} 
}
\email{
    ehtesham@amazon.com
}

\author{
    Purak Jain
}
\affiliation{
    \institution{Amazon Inc.}
    \country{} 
}
\email{
    purakjn@amazon.com
}

\usepackage[utf8]{inputenc} 
\usepackage[T1]{fontenc}    
\usepackage{hyperref}       
\usepackage{url}            
\usepackage{booktabs}       
\usepackage{nicefrac}       
\usepackage{microtype}      
\usepackage{xcolor}         
\usepackage{graphicx}
\usepackage{tabularx}
\usepackage{multirow}
\usepackage{float}
\usepackage{graphicx}
\usepackage{algorithm}
\usepackage{algpseudocode}
\usepackage{wrapfig}
\usepackage{fancyvrb}
\usepackage{fancyvrb}
\usepackage{fvextra}
\usepackage{makecell}

\usepackage{adjustbox}
\usepackage{natbib}
\usepackage{caption}

\usepackage{pgfplots}
\usepackage{pgfplotstable}
\usepackage{booktabs}
\pgfplotsset{compat=1.17}
\usepackage{tikz}
\usepackage{pgf-pie}
\usepackage{caption}
\usepackage{subcaption}

\usepackage{pifont}
\begin{document}

\title{LLMs for Customized Marketing Content Generation \\
and Evaluation at Scale} 



\begin{abstract}
Offsite marketing is essential in e-commerce, enabling businesses to reach customers through external platforms and drive traffic to retail websites. However, most current offsite marketing content is overly generic, template-based, and poorly aligned with landing pages, limiting its effectiveness.
To address these limitations, we propose MarketingFM, a retrieval-augmented marketing content generation system that integrates multiple data sources to produce keyword-specific ad copy with minimal human intervention. We validate MarketingFM through offline human and automated evaluations and large-scale online A/B tests. In a recent experiment, keyword-focused ad copy outperformed template-based ads, achieving up to 9\% higher click-through rates (CTR), 12\% more impressions, and a 0.38\% lower cost-per-click (CPC), demonstrating improved ad ranking and cost efficiency.
Despite these gains, manual review of generated ads remains costly and time-consuming. To mitigate this, we introduce AutoEval-Main, an automated evaluation system that combines rule-based metrics and LLM-as-a-Judge approaches, ensuring alignment with marketing principles. 
In experiments conducted with large-scale high-quality human annotation data, AutoEval-Main achieves an agreement rate of 89.57\% with human reviewers.
Building on this, we further propose AutoEval-Update, a cost-efficient LLM-human collaborative framework designed to dynamically refine the evaluation prompt and adapt to shifting criteria with minimal human effort. 
By selectively sampling representative ads for human review and leveraging a critic LLM to generate alignment reports, AutoEval-Update enhances evaluation consistency while significantly reducing manual effort.
Our experiments show that the critic LLM proposes meaningful refinements, enhancing alignment between LLM-based and human evaluations.
However, human oversight remains crucial for setting thresholds and validating refinements before full-scale deployment.

\end{abstract}


\received{20 February 2007}
\received[revised]{12 March 2009}
\received[accepted]{5 June 2009}

\maketitle

\section{Introduction}

\begin{figure}[tbp]
    \centering
    \vspace{10 pt}
    \includegraphics[width=0.48\textwidth]{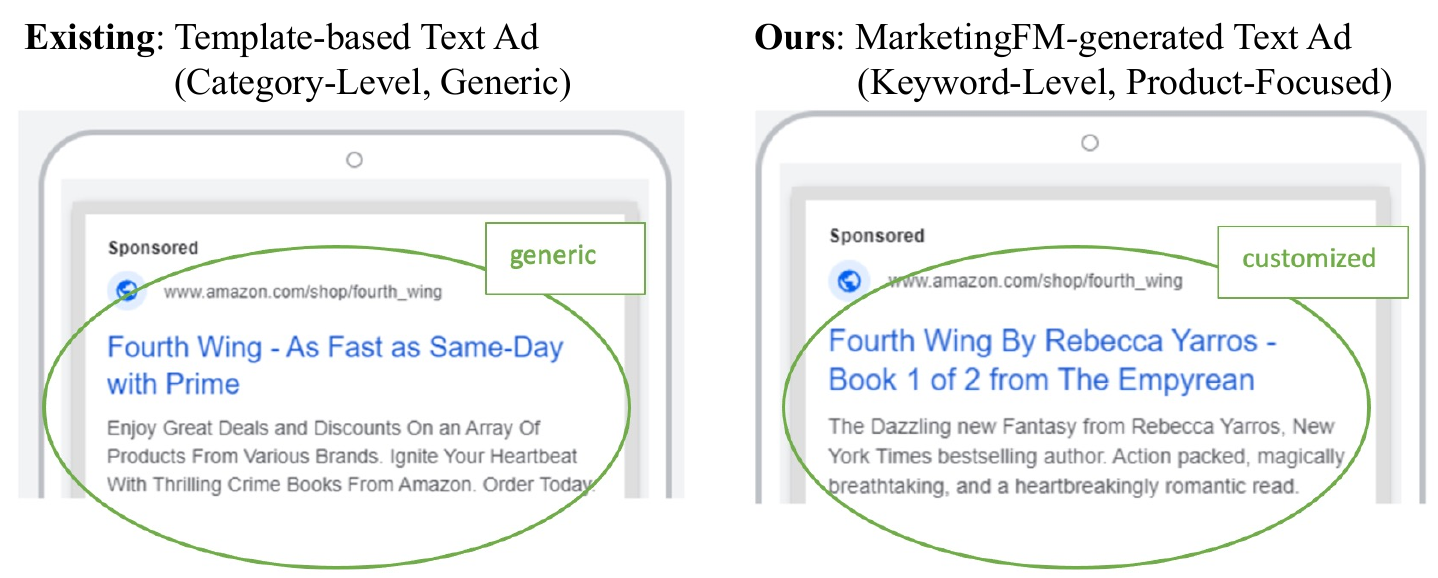}
    \vspace{-10 pt}
    \caption{A comparison of two paid search ads for an e-commerce website targeting the same keyword. The left ad is generic and template-based, while the right is our keyword-specific, product-focused approach to enhance ad relevance and user engagement considering product context. 
}
    \label{fig:ps-example}   
    \vspace{-5pt}
\end{figure}



Customers often search for products through web services like Google, Bing, and social media services such as Facebook, Instagram, and Pinterest, where they encounter e-commerce advertisements alongside organic search results. To effectively engage users and present the products of an e-commerce website, these \textit{offsite marketing content} must be carefully crafted to clearly communicate product offerings and present them in an appealing way.

Currently, most large-scale e-commerce websites rely on manually curated, \textit{template-based} ad copies, images, and videos that fail to adapt to their diverse product offerings. 
In search marketing, marketers often craft broad category-level templates like "Come Find Your Favorite <Keyword> on [e-commerce website]!" for entire keyword categories without emphasizing product-specific features. While this approach provides a structured framework, it suffers from several key limitations. First, generic templates fail to capture product-specific nuances that could differentiate offerings in a competitive marketplace. 
Second, they struggle to adapt messaging to varying customer intents across different channels, such as search and social marketing channels. 
Third, the resource-intensive nature of manual content creation hinders scalability, making it difficult to customize or personalize content at scale. 
As a result, traditional methods often misalign ad content with user expectations, leading to limited audience engagement and suboptimal returns on advertising investment.

\begin{figure*}[tbp]
    \centering
    \vspace{-5pt}
    \includegraphics[width=0.8\textwidth]{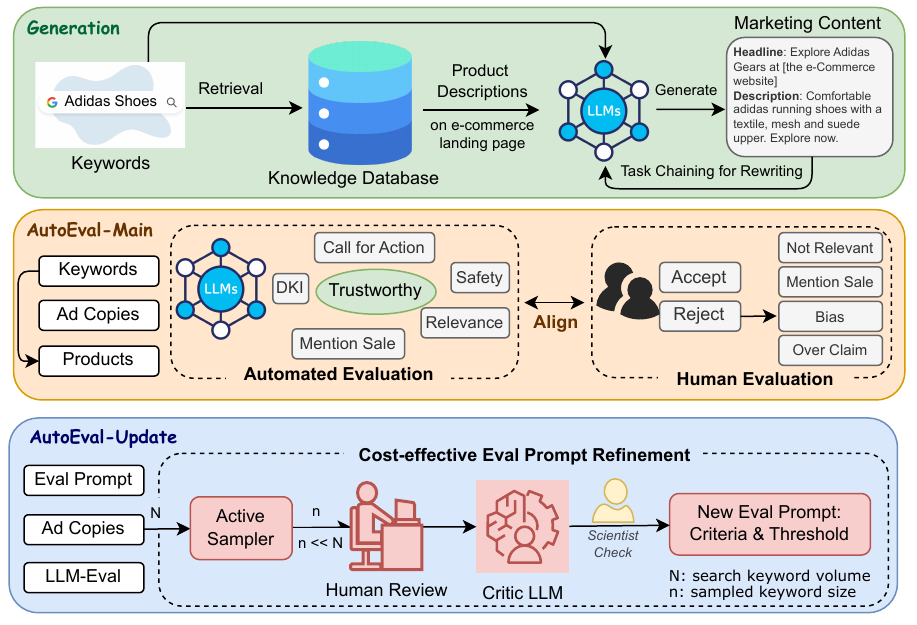}
    \vspace{-10pt}
    \caption{An illustration of the proposed MarketingFM framework for marketing content generation and evaluation.
    The top section represents the ad copy generation pipeline, where retrieval-augmented generation and task chaining ensure relevant and concise ad content.
    The middle section depicts AutoEval-Main, an automated evaluation system that assesses ad quality based on key criteria, aligning LLM-generated content with human evaluations.
    The bottom section illustrates AutoEval-Update, a cost-effective LLM-human collaborative evaluation refinement process that iteratively updates evaluation criteria.
    }
    \label{fig:framework}
\end{figure*}

Recent advances in Large Language Models (LLMs)~\cite{brown2020language, chowdhery2023palm, touvron2023llama, yang2024kg, dong2024survey} offer promising opportunities for automating marketing content generation at scale \cite{chung2023increasing, wu2025survey, 10.1145/3604915.3610647, bismay2024reasoningrec}. 
However, directly applying these models introduces risks such as factual inaccuracies~\cite{celikyilmaz2020evaluation, chen2025graphcheck, liang2024controllable}, unsubstantiated claims~\cite{qiu2023detecting, farquhar2024detecting}, and potential policy violations~\cite{karamolegkou2023copyright, neel2023privacy, dong2024disclosure}, all of which can degrade customer experience and engagement. 
For example, for the keyword ``Stanley Cup'', a baseline model mistakenly generated hockey-related ad copy instead of content for trending water bottles, underscoring the importance of aligning generated content with both customer intent and product catalog accuracy.

To address these challenges and enhance ad copy customization without intensive human labor, we introduce \textbf{MarketingFM}, a framework that automates and optimizes the generation of marketing content at scale. MarketingFM empowers e-commerce marketing teams to generate custom ad content for millions of products while ensuring relevance to diverse customer interests. 
By using Retrieval-Augmented Generation (RAG)~\cite{lewis2021retrievalaugmented, gao2023retrieval, kortukov2024studying} with existing onsite search infrastructure within the e-commerce service, MarketingFM grounds ad generation in real-time product data corresponding to customer queries, thereby enhancing ad relevance.
Figure~\ref{fig:ps-example} illustrates how MarketingFM produces more relevant and targeted ad content compared to generic ad copy in the paid search use case.
To assess the effectiveness of MarketingFM's ad generation, we conducted large-scale human evaluations across multiple criteria, demonstrating a high acceptance rate for the generated ads. Additionally, online A/B testing showed a significant increase in user engagement, confirming the framework’s ability to produce high-quality, relevant ad content at scale.

To further reduce the need for costly human review, we developed \textbf{AutoEval-Main}, an automated evaluation system that integrates \textbf{rule-based checks} and \textbf{LLM-as-a-Judge scoring} to ensure ad quality. Rule-based methods enforce policy compliance by detecting disallowed terms, formatting issues, and stylistic inconsistencies, ensuring adherence to channel-specific requirements. Meanwhile, the LLM-as-a-Judge component evaluates query relevance and factual accuracy by cross-referencing product metadata, preventing hallucinations, overclaims, and policy violations. By leveraging human evaluation data as a benchmark, we demonstrate that this hybrid approach closely aligns with human judgments, maintaining high-quality standards while improving efficiency in content validation.

While AutoEval-Main enhances efficiency and consistency, it requires dynamic refinement to keep pace with evolving customer behavior, keyword distributions, and evaluation standards. 
To address this criteria drift issue, we propose \textbf{AutoEval-Update}, an automated self-refinement framework that iteratively improves the evaluation prompts and criteria. This cost-effective solution leverages \textit{active sampling} and a \textit{critic LLM} to identify misalignment factors and dynamically refine evaluation standards.
The process begins by actively sampling a small but representative subset of ad copies for evaluation by AutoEval-Main and human annotators. The critic LLM, a more advanced model than the LLM of ad generation, then generates an alignment report by identifying discrepancies between human evaluations and AutoEval-Main outputs, derived from new evaluation criteria. 
These refinements are iteratively incorporated into AutoEval, ensuring that AutoEval remains adaptive and aligned with emerging needs.
This process continues until convergence, with only the final round of updates undergoing domain expert review before deployment to ensure robustness and accuracy in large-scale ad evaluation. 
Experiments confirm that AutoEval-Update enables refinement in evaluation quality with little human labor, ensuring sustained alignment with human assessments and adaptability to evolving ad evaluation standards.

In summary, our contributions are:
\begin{itemize}
   \item We propose \textbf{MarketingFM} that ensures LLM-generated ad copies align with marketing principles. We curate ad quality criteria based on marketer insights and benchmark this task using large-scale human annotations.

    \item We develop \textbf{AutoEval-Main}, an automated evaluation system that integrates rule-based checks and LLM-as-a-Judge to assess ad quality, reducing need for costly human review.

    \item We introduce \textbf{AutoEval-Update}, a self-refining evaluation framework that actively samples data and automatically updates evaluation criteria to adapt to shifting data and evolving standards, ensuring continuous improvement of the automated evaluation system.

    \item We validate the effectiveness of our systems through both \textbf{offline evaluation} and \textbf{online A/B tests}, demonstrating improvements in marketing content quality and alignment with marketer objectives.

\end{itemize}

Notably, our MarketingFM framework can be extended to other marketing use cases, such as social media and outbound marketing email, demonstrating the potential of foundation models for marketing content generation. 

\section{Related Work}

\paragraph{LLMs for Ads Generation.}
Relevant ad copy is crucial for driving engagement and optimizing the performance of sponsored links. Limited prior works have explored the role of LLMs in ad generation, with a survey paper~\cite{reisenbichler2023applying} highlighting their potential in search engine advertising and emphasizing the importance of structured guidance to ensure quality and relevance. Likewise, another study~\cite{zhou2024gcof} improves CTR by integrating LLMs with feature engineering and modified genetic algorithms for copywriting, demonstrating the need to align ad content with user intent, landing pages, and performance metrics.
In contrast, our work consider practical deployment of LLMs for offsite search marketing in e-commerce, integrating marketing expertise into ad generation. We design domain-specialized prompts and RAG mechanism to drive engagement, integrate task chaining~\cite{huang2023pcr, wei2022chain} to meet character limits, and establish a dataset with high-quality, large-scale human annotations to evaluate LLM-generated ad quality.

\paragraph{LLM-as-a-Judge.}
The evaluation of language models, particularly in content generation, necessitates robust and reliable assessment methods. Recent studies have underscored the effectiveness of using foundation models with shared datasets to ensure consistent evaluations and minimize bias~\cite{zheng2023judging, gu2024survey}. Metrics such as fluency, coherence, relevance, and n-gram diversity have been widely adopted to evaluate model outputs~\cite{wang2023decodingtrust}. Additionally, researchers have investigated using strong LLMs as evaluators to reduce the costs associated with human assessment. These LLM-based evaluations have proven effective in detecting biases, such as position and verbosity bias, achieving over 80\% agreement with human judgments on benchmark datasets, thus demonstrating the scalability and cost-efficiency of LLMs as evaluative tools~\cite{yasunaga2023retrievalaugmented}.
Unlike prior work on general-domain question answering~\cite{liu2023g, li2024iqa}, reasoning~\cite{setlur2024rewarding, saha2025learning, wang2023can, zhou2024self}, and alignment~\cite{ashktorab2024aligning, liang2024sheep, zhang2024self}, we adapt LLM-as-a-Judge for marketing content quality evaluation aligning with marketer principles. Our system evaluates the relevance of the ad copy and the generalizability of the ad, ensuring relevance with search keywords and products from the landing page. To enable scalability, we design it for both effectiveness and efficiency in large-scale advertising applications.

However, in real-world applications, evaluation criteria evolve as models generate new outputs and external conditions shift.
\citet{shankar2024validates} highlights this \textit{criteria drift} problem—users need criteria to grade outputs, yet grading outputs help define those criteria. Their work proposes a UI-based human-LLM collaboration system, but it still depends on manual updates, making the process labor-intensive and difficult to scale.
In contrast, we introduce an automated framework designed to meet industry-scale evaluation needs by minimizing human effort. Using active sampling and a powerful critic LLM, our system ensures continuous adaptation to shifting customer interests, evolving product catalogs, and changing advertising requirements. 
Our approach enables scalable and high-quality ad evaluation in dynamic commercial environments.



\section{LLM for Ads Content Generation}
\label{headings}

\paragraph{Problem Setting.}
Given an input customer intent query $x$ (the search keyword), we retrieve relevant product documents $z$, which provide titles and detailed product features as context. This context, along with $x$, is used to generate marketing content $y$. 
For example, in the Paid Search use case, the output $y$ is the ad \textit{titles} and \textit{descriptions}, as shown in Figure~\ref{fig:ps-example}. 
Formally, our generation process can be represented as:
\begin{equation*}
    p_{\text{MarketingFM}}(y|x) = p_{\theta}(y|x, z) p_{\phi}(z|x)
\end{equation*}
Specifically, the generation process has two components:
(1) a retriever $p_\phi$ that retrieves the top $k$ product documents $z$, conditioned on $x$;
(2) a generator $p_\theta$ that produces $y$ based on $x$ and the retrieved context $z$.
An illustration of the marketingFM framework is shown in Figure~\ref{fig:framework}.

\subsection{Search Keyword Context Retrieval} 
To resolve ambiguity and improve ad relevance, we retrieve products from a knowledge base (KB) of all products from an e-commerce website and use the product metadata as context for generation.
For this RAG component, we propose two retrieval strategies: (1) semantic embedding-based retrieval and (2) search page products context retrieval.

\paragraph{Semantic Embedding-based Retrieval.}
Suppose that, given the input $x$, $z^+$ is a target product to retrieve \footnote{For simplicity, we consider one $z^+$ for
each input; the retrieval target can be a set of triplets $\{z^+\}$}. Then, the objective is formulated with a sentence encoder model $p_\phi$ as:
\begin{equation*}
    z^+ = \text{argmax}_{z\in \text{KB}} f(p_\phi(x), p_\phi(z))
\end{equation*}
where $f$ is a non-parametric scoring function that calculates the similarity between the input query representation
$p_\phi(x)$ and document embedding $p_\phi(z)$, for example, the dot product.
Specifically, we use all-MiniLM-L6-v2~\footnote{\url{https://huggingface.co/sentence-transformers/all-MiniLM-L6-v2}} model~\cite{huggingface2020allminilml6v2}.
To efficiently retrieve the top-$k$ documents from millions of high-dimensional vectors, we use the FAISS library~\cite{douze2024faiss} for document indexing and similarity calculation.

\paragraph{RAG using Search Page Products Context.}
In this approach, we use a different KB that directly maps search keywords to products from the e-commerce website’s search result pages, assuming their relevance for ad copy generation.
Specifically, we prioritize products from earlier search result pages and select the top-$k$ products to extract metadata, \textit{i.e.}, product title and descriptions. This metadata forms the context for each search keyword. Since the products displayed on search pages may change over time, we regularly refresh the KB to ensure accuracy.

\subsection{Marketing Content Generation and Task Chaining} \label{method: gen}
For content generation $p_\theta(y|x, z)$, we use Anthropic Claude Haiku~\cite{anthropic2024claude} through Amazon Bedrock~\footnote{\url{https://aws.amazon.com/bedrock/}} to balance the cost of API and the quality of the generation. 
Specifically, for a given \textbf{\textit{search keyword}} $x$ for advertising, we construct a \textbf{\textit{prompt}} $P$ by retrieving the \textit{\textbf{context}} $z$ and then use $P$ as input to $p_\theta$. In practice, based on suggestions from marketing managers, we design two prompts: a general-purpose prompt and a call-to-action prompt (intended to encourage customer engagement, definitions in Appendix~\ref{appendix: marketer_insight}), and we use both in combination (implementation details in Section~\ref{sec: gen-offline-eval}).

\noindent\paragraph{Task Chaining.}
However, ensuring rule-following remains a challenge for LLMs~\cite{mu2023can, sun2024beyond}, and one key requirement for paid search ads is to keep headlines under 30 characters.
This is difficult due to the length of many keywords (over 50\% exceeding 15 characters). Despite trying various prompts, our initial attempts failed as most generated headlines exceeded the limit. 
To resolve this, we adopt a \textit{task chaining} approach through \textit{summarization}: First, we generate twice as many headlines and descriptions as needed and then use a subsequent LLM call to summarize them so they meet the length requirement. This approach effectively produces headlines within the limit using just two LLM calls, whereas the previous method could require up to 10 attempts to satisfy length constraints.
We provide the prompt template in Appendix~\ref{appendix: prompt-rewrite}.

\section{Benchmarking LLM-Generated Ads at Scale: Quality Standards and Human Evaluation}
We collaborate with marketers to define detailed ad quality requirements and create guidelines for human annotators, ensuring alignment with marketing objectives and customer engagement goals. The requirements for engagement-driven messaging and keyword relevance are detailed in Appendix~\ref{appendix: evaluation_prompt}, while Table~\ref{tab: human_rejection_reasons} shows rejection reasons and theirdefinitions.
Additionally, we develop a human annotation workflow tool using AWS GroundTruth to support human evaluation. 
The annotators review all headlines and descriptions for each keyword, accepting or rejecting content based on guidelines provided by business stakeholders and the marketing team.
The process provides an \textit{accept} or \textit{reject} status for each ad copy, along with the specific \textit{reason for rejection}. 
As shown in Table~\ref{tab:human_eval_overall}, over 97\% of generated ad copies are accepted, indicating high-quality LLM generation. 
A breakdown of the rejected reasons are in Table~\ref{tab:human_eval_reasons}, with ``not relevant'' being the most common reason for rejection. 
Note that the rejection reason \textit{bias} does not indicate harmful stereotyping but rather cases where an ad copy \textit{unintentionally limits the target audience} based on gender, age, or other demographics. For example, if a user searches for ``birthday gifts'' but the ad copy states ``birthday gifts for kids'', it narrows the intended audience. While this does not pose a fairness risk, it may affect inclusivity and ad performance.  
Notably, \textbf{no safety or toxicity issues} are identified in the generated content, confirming the trustworthiness of our system.


\paragraph{Analysis on Human Evaluation Quality.}
To assess reviewer reliability, we randomly sample and manually validate the annotations. Due to the data imbalance (only 2\% of the cases are rejections), we calculated weighted metrics to reflect the proportion of accepted and rejected cases. As shown in Table~\ref{tab:val_human_eval}, the overall weighted error rate is 3.71\%, confirming the trustworthiness of human evaluations.

With these human annotations, we establish a \textbf{comprehensive benchmark dataset} for ad copy generation, encompassing keywords, product metadata, generated ad copies, and human labels. The dataset enables us to (1) validate our offline evaluation metrics against human judgment while (2) providing valuable training data for future model fine-tuning.

\begin{table}[tbp]
    \centering
    \begin{subtable}[t]{0.2\textwidth}
        \centering
        \resizebox{\textwidth}{!}{
        \begin{tabular}{lc}\toprule
        \textbf{Eval. Results} &\textbf{Ratio (\%)} \\\midrule
        accept &97.21 \\
        reject &2.79 \\
        \textbf{total} &100 \\
        \bottomrule
        \end{tabular}}
        \caption{Overall human evaluation results.}
        \label{tab:human_eval_overall}
    \end{subtable}
    \hfill
    \begin{subtable}[t]{0.2\textwidth}
        \centering
        \resizebox{\textwidth}{!}{
        \begin{tabular}{lc}\toprule
        \textbf{Rej. Reason} &\textbf{Ratio (\%)} \\\midrule
        not relevant &48.64 \\
        too specific &22.49 \\
        over claim &13.92 \\
        mention sales &7.11 \\
        bias &5.96 \\
        too creative &1.75 \\
        others &0.14 \\
        \textbf{total} &100 \\
        \bottomrule
        \end{tabular}}
        \caption{Breakdown of rejections.}
        \label{tab:human_eval_reasons}
    \end{subtable}
    \caption{Human evaluation results for 150,000 generated ad copies of 10,000 keywords.}
    \label{tab:val_human_eval}
    \vspace{-20pt}
\end{table}

\begin{table}[tbp]
\centering
\label{tab:human-review}
\resizebox{0.4\textwidth}{!}{
\scriptsize
\begin{tabular}{lcc}
\toprule
\textbf{Human Opinion} & \textbf{Correct (\%)} & \textbf{Mistake (\%)} \\
\midrule
\textbf{Accept} & 97 & 3 \\
\textbf{Reject} & 71 & 29 \\
\midrule
\textbf{Weighted Accuracy} & \textbf{96.29} & \textbf{3.71} \\
\bottomrule
\end{tabular}}
\caption{Sampled and weighted manual validation results for human annotations.}
\vspace{-20pt}
\end{table}

\begin{table*}[tbp]
\centering
\resizebox{0.9\textwidth}{!}{
\renewcommand{\arraystretch}{1.0} 
\begin{tabular}{l p{10cm} c c}
\toprule
\textbf{Criterion} & \textbf{Description} & \textbf{Evaluation Method} & \textbf{Scoring} \\
\midrule
Relevance & Ad copy should balance between vague and specific, align with keyword/search intent, avoid bias, and not focus narrowly on one or two products. & LLM-as-a-Judge & 0–5 Score \\ \midrule
Generalization & Evaluates how well the ad copy generalizes while remaining relevant, avoiding excessive specificity or bias. & LLM-as-a-Judge & 0–5 Score \\\midrule
Safety & Content must be appropriate, avoiding harmful/offensive language and adhering to advertising standards. & Rule-based & Accept/Reject \\\midrule
Mention Sale & Ad copy should avoid misleading price impressions. No "sale," "discounts," or "clearance" are allowed. & Rule-based & Accept/Reject \\\midrule
Diversity & For a keyword, ad copies should vary in tone, features, or aspects to ensure diversity compared to other ads for the same keyword. & Rule-based & Accept/Reject \\
\bottomrule
\end{tabular}}
\caption{AutoEval-Main Criteria: Our proposed evaluation system criteria for assessing the quality of generated ad copies. The framework combines LLM-based evaluation and rule-based checks, with scoring formats varying between numerical ratings and binary accept/reject decisions.} 
\label{tab:criteria_ad_copy}
\vspace{-15pt}
\end{table*}

\section{Automated Evaluation for Generated Ads}
To reduce the cost associated with human annotation and screening, we explore automated evaluation methods using LLMs. 

\subsection{AutoEval-Main: Evaluation via LLM-as-a-Judge and Rule-Based Checks}
Our proposed automated evaluation framework, AutoEval-Main (Figure \ref{fig:framework}.middle), integrates Rule-Based Checks with an LLM-as-a-Judge system \cite{zheng2023judging} to ensure the generation of high-quality ad copies. 
The Rule-Based Checks implement efficient, stoplist-based methods that address inherent limitations in LLM outputs to ensure compliance with business requirements by accepting/rejecting ad copies based on criteria defined in Table \ref{tab:criteria_ad_copy}. 
The LLM-as-a-Judge system introduces a more nuanced evaluation layer by assessing ad copies on two key metrics: Relevance and Generalization (refer to Table \ref{tab:criteria_ad_copy}), each rated on a 0-5 scale. 
The system leverages prompt engineering (refer to Appendix~\ref{appendix: evaluation_prompt} for detail) and domain expert-crafted few-shot examples, complemented by query-specific context from the KB. These components guide the LLM in evaluating ad copies against predetermined criteria, with the examples serving as evaluation benchmarks.


The workflow begins when a user submits a search query, e.g., ``wireless headphones for gym'', prompting the system to generate multiple ad variations, typically 12 headlines and 3 descriptions. These ad copies first undergo rule-based checks to filter out those that fail to meet fundamental requirements. The remaining copies, along with the original query and relevant product context from the KB, are then evaluated by the LLM. The system assigns Relevance and Generalization scores to each ad copy, accepting those that meet or exceed the threshold while rejecting the rest.


\setlength{\textfloatsep}{10pt}
\begin{algorithm}[tbp]
\caption{AutoEval-Update Framework for Self-Refining Ad Copy Evaluation}\label{alg:auto_eval_update}
\begin{algorithmic}[1]

\Require
\State LLM-generated ad copies $C$
\State Initial evaluation prompt $P$
\State Evaluation criteria and KB context
\State Performance threshold $\tau$
\State Maximum iteration rounds $N$

\Ensure
\State Refined evaluation prompt $P'$ aligned with human judgment.

\State \textbf{Initialization:} Select a diverse subset $C_s \subseteq C$ for evaluation.

\State \textbf{Step 1 - Active Sampling:} 
Ensure $C_s$ represents varied query types, product categories, and prior evaluation outcomes.

\State \textbf{Step 2 - Automated Evaluation:} 
Use prompt $P$ to score $C_s$, generating LLM evaluation scores $S_s$.

\State \textbf{Step 3 - Human Evaluation:}  
\begin{itemize}
    \item Obtain human feedback $H$ on $C_s$.
    \item Identify discrepancies between $S_s$ and $H$.
\end{itemize}

\State \textbf{Step 4 - Discrepancy Analysis:}  
\begin{itemize}
    \item Compare $S_s$ and $H$ to detect misalignments.
    \item Diagnose root causes (e.g., ambiguous criteria, LLM biases).
\end{itemize}

\State \textbf{Step 5 - Prompt Refinement:}  
\begin{itemize}
    \item Update $P$ based on insights from Step 4, generating refined prompt $P'$.  
    \item Adjust evaluation guidelines to enhance clarity and penalize inconsistencies.  
\end{itemize}

\State \textbf{Iteration:}
Apply $P'$ in subsequent evaluations and repeat Steps 1–5 iteratively. Iteration terminates when evaluation performance meets a predefined threshold $\tau$ or exhibits no significant improvement over $N$ consecutive rounds.

\State \textbf{Output:} Refined evaluation prompt $P'$.

\end{algorithmic}
\end{algorithm}

\subsection{AutoEval-Update: Iterative Self-Refinement with LLM-Human Collaboration}

Ad copy evaluation is inherently dynamic, shaped by data distribution shifts, evolving marketing priorities, and policy changes. These factors necessitate regular updates to evaluation prompts to maintain relevance and accuracy.
For instance, growing public concern over \textit{product safety} may necessitate more rigorous accuracy metrics, while increasing scrutiny on \textit{political or socially sensitive advertising} could require more precise or inclusive language guidelines to ensure compliance with evolving regulations.

The proposed AutoEval-Update framework (Figure~\ref{fig:framework}.lower) addresses these challenges through an iterative refinement process that continuously improves evaluation prompts. 
AutoEval-Update begins with active sampling, selecting a diverse and representative subset of LLM-generated ad copies and their evaluation results. 
These samples undergo human review to validate alignment with LLM assessments, establishing a reliable ground truth. 
Discrepancies between human and LLM evaluations are analyzed in an alignment report, identifying issues such as keyword misinterpretation or inconsistent criteria application. 
Insights from this analysis guide prompt adjustments, refining evaluation criteria and thresholds for the next iteration. 
The process continues until evaluation performance meets a predefined threshold $\tau$ or shows no significant improvement over $N$ consecutive rounds. 
At scale, this process ensures that evaluation remains adaptive and aligned with evolving data with little human involve.
For instance, from a daily digest of 20 million keywords, the active sampler selects 1,000 keywords with their generated ads for human review, which, along with the critic LLM’s feedback, iteratively refines the evaluation prompt before applying it to new assessments.
The detailed procedure is outlined in Algorithm~\ref{alg:auto_eval_update}, and Figure~\ref{fig:autoeval-update-framework} illustrates the end-to-end production pipeline.

\paragraph{Active Sampling Strategies.}
Selecting a representative and meaningful subset for human review is crucial for maintaining evaluation quality. To achieve this, the framework employs three sampling approaches:
(1) \textit{Random Sampling}: A baseline method that selects samples uniformly at random. While simple and unbiased, it may fail to capture critical edge cases;
(2) \textit{Uncertainty-Based Sampling}: Leverages LLM-generated confidence scores to prioritize samples with lower confidence, ensuring that cases with ambiguous or uncertain evaluations receive targeted human review.
(3) \textit{Diversity-Based Sampling}: Ensures a broad and balanced representation by selecting samples across different categories, CTR segments, or click distributions, capturing variations in data distribution and model performance.

\section{Experiments and Lessons Learned}
We select 10,000 high-performing keywords for paid search ad copy generation, with a CTR range of 5\% to 60\%. Using our framework with Anthropic Claude as the base model, we generate ad copies for these keywords. We present the results from both offline evaluations and online tests, along with key insights and findings.

\subsection{Offline Evaluation of Generated Marketing Content} \label{sec: gen-offline-eval}
Given the public-facing nature of our offsite outputs, even a few low-quality ads could damage the brand’s reputation and adversely affect the e-commerce site’s business. Therefore, ensuring the quality and safety of every output is paramount. 
To mitigate these risks and maintain high standards, we conducted rigorous offline evaluations—combining human assessment with LLM-as-a-Judge—before launching online A/B tests.


\begin{table}[tbp]
    \centering
    \resizebox{0.28\textwidth}{!}{
    \begin{tabular}{lc}
        \toprule
        \textbf{Relevance Score} & \textbf{Percentage} \\
        \midrule
        High Relevant (5)      & 54.34\% \\
        Relevant (4)           & 20.49\% \\
        Moderately Relevant (3) & 19.09\% \\
        Slightly Relevant (2)  & 5.52\% \\
        Not Relevant (1)       & 0.56\% \\
        \bottomrule
    \end{tabular}}
    \caption{Distribution of LLM-evaluated relevance scores}
    \label{tab: relevance_scores}
    \vspace{-10pt}
\end{table}

\paragraph{Implemention Details.}
During MarketingFM's generation process, we generate 
then filter and re-generate ad copies based on three key marketer requirements: \textit{relevance} to landing page product context, \textit{call-to-action} (CTA), and \textit{dynamic keyword insertion} (DKI) (definitions in Appendix~\ref{appendix: marketer_insight}).
For relevance, we provide LLM-generated headlines to the critic LLM, requesting it to score the relevance to the search keyword on a scale from 0 to 5, where 0 indicates no relevance and 5 indicates high relevance. 
For this experiment, any ad copy with a score less than 4 was automatically rejected and replaced with a new headline. 
For CTA, we construct a list and check the presence of action-oriented verbs such as ``Buy Now'' or ``Shop at [the e-commerce website]''. Headlines lacking CTA verbs are removed and regenerated. 
For DKI, we do not enforce strict rules due to keyword length variability and character limits imposed by the search engine advertising service. Instead, we analyze the output to assess keyword incorporation distribution, enabling automatic filtering and refinement of ad copies based on key marketing criteria.

\paragraph{Automated Evaluation Results.}
Table~\ref{tab: relevance_scores} shows the relevance scores for raw (unfiltered) ad copies, with 74.83\% scoring 4 or higher (regeneration threshold). CTA and DKI analysis is in Appendix~\ref{appendix:cta_dki}.
Notably, we observe that LLM-based relevance scoring lacks precision at mid-scale (2–4), making it difficult to establish a concrete irrelevance threshold. This misalignment with marketer-defined rejection criteria reveals the limitations of manually-defined evaluation criteria and threshold optimization. These insights led us to further develop AutoEval-Update, enabling dynamic refinement of evaluation criteria and thresholds.

\begin{table}[tbp]
    \centering
    \vspace{-5pt}
    \resizebox{0.4\textwidth}{!}{
    \begin{tabular}{lcc}
        \toprule
        \textbf{Category} & \textbf{Percentage} \\
        \midrule
        Overall Agreement  & 89.57\% \\
        Overall Disagreement & 10.43\% \\
        \midrule
        \textit{Breakdown of Disagreements} & & \\
        \quad Human Accepted but LLM Rejected & 8.90\% \\
        \quad Human Rejected but LLM Accepted & 1.53\% \\
        \bottomrule
    \end{tabular}}
    \caption{Alignment analysis between LLM-based and human evaluations.}
    \label{tab:agreement_analysis}
    \vspace{-20 pt}
\end{table}


\subsection{Online Test for Generated Marketing Content} 


The hypothesis driving this initiative posits that keyword-specific ad copy boosts engagement among shoppers on our partner search engine advertising service. To validate this, we conducted a randomized A/B test across 10,000 keywords with a wide range of CTRs. More details on the experimentation infrastructure that enables scalable experiment splits and execution can be found in~\citet{jain2024serp}. In the experiment, the Control group used generic or category-level ad copy, while the Treatment group utilized keyword-specific copy.
We begin with a 3,000-keyword online test using the semantic-based retrieval method. The results, shown in Table~\ref{tab: eval-online-3k}, measure engagement using click-through-rate (CTR), defined as clicks over impressions. The treatment group shows an increase in CTR, particularly on mobile devices, which represent 80\% of our traffic.

\begin{table}[tbp]
\centering
\resizebox{0.23\textwidth}{!}{
\begin{tabular}{lrrrr}
\toprule
\textbf{Device} & \textbf{Lift (bps)} & \textbf{p-value} \\
\midrule
Mobile  & \textbf{+121} & 0.055 \\
Desktop & \textbf{+38} & 0.269 \\
\bottomrule
\end{tabular}}
\caption{First-round online A/B test results (CTR) comparing generic ad copy (control) to keyword-specific ad copy (treatment) for 3,000 keywords on a major search engine ads service. The retriever used in RAG is semantic-based.} 
\label{tab: eval-online-3k}
\vspace{-10pt}
\end{table}

\begin{table}
\centering
\resizebox{0.4\textwidth}{!}{
\begin{tabular}{lcc|cc}
\toprule
\textbf{Metric} & \multicolumn{2}{c|}{\textbf{Mobile}} & \multicolumn{2}{c}{\textbf{Desktop}} \\
 & \textbf{Lift} & \textbf{p-value} & \textbf{Lift} & \textbf{p-value} \\
\midrule
Clicks Lift & \textbf{+8\%} & 3.89e-6 & \textbf{+9\%} & 2.56e-3 \\
Impressions Lift & \textbf{+8\%} & 3.17e-6 & \textbf{+12\%} & 1.01e-5 \\
CTR Lift (bps) & \textbf{(+4)} & 2.34e-1 & \textbf{(+24)} & 1.10e-1 \\
CPC Reduction & \textbf{-0.35\%} & 3.07e-3 & \textbf{-0.22\%} & 2.01e-1 \\
\bottomrule
\end{tabular}}
\caption{Second-round online A/B test results (CTR, clicks, CPC) comparing keyword-specific ad copy (treatment) to template-based ad copy (control) for 10,000 keywords on a major search engine advertising service. The retrieval method is products context KB-based RAG. 
}
\label{tab:eval-online-10k}
\vspace{-10pt}
\end{table}

From the test, we identified a relevancy issue of 10\%, where the ad copy did not match the products on the e-commerce website’s search landing pages. 
To address this, we curate a KB mapping keywords to search page products, which reduce the relevancy issue to 1\%. Furthermore, our partner search engine advertising service recommended using a specific output style that incorporates CTAs and DKIs (definitions in Appendix~\ref{appendix: marketer_insight}). The post-experiment analysis also confirmed that successful keyword segments contain CTA and DKI. 
As a result, we implemented these guidelines through task chaining.
Building on this, we expand to a 10,000-keyword online test (Table~\ref{tab:eval-online-10k}), where keyword-specific Ad Copy (Treatment) achieves an \textbf{8\% increase in impressions on mobile and a 12\% increase on desktop}, indicating \textbf{improved visibility and ad rank performance on the tested major search engine advertising service}. While CTR remains stable due to broader reach, overall click volume increases, and the lower CPC indicates improved cost efficiency. 
We observe that aligning LLM output with the advertising service provider's preferred style and tone is key to increasing impression share and search ranking.

\subsection{Evaluation of AutoEval-Main} \label{sec: autoeval-main-eval}
The effectiveness of an evaluation framework for LLM-generated ad copies depends on balancing \textbf{concerned} rate (incorrectly approving unacceptable content, LLM accept but human reject rate) and \textbf{wasted} rate (rejecting acceptable content, LLM reject but human accept rate). 
To evaluate the AutoEval framework, we tested its performance on 150,000 ad copies across 10,000 keywords, using a human-labeled dataset as the benchmark. Each query included 12 headlines and 3 descriptions generated by the LLM. Human reviewers assessed the same ad copies, assigning Accept or Reject labels based on predefined criteria outlined in Table~\ref{tab:criteria_ad_copy}.

With offline evaluation with AutoEval-Main, we find LLM and human evaluations agree with each other in 89.57\% of cases, indicating substantial alignment (Figure~\ref{tab:agreement_analysis}). Notably, we notice from outputs that LLM tends to be more conservative than human evaluators, often rejecting ad copies that humans accept. This suggests that calibrating LLM's evaluation criteria could enhance alignment with human perspectives and reduce unnecessary rejections of acceptable content.

An ablation study on three LLM-as-a-Judge modes, applying 0–5 thresholds for relevance and generalization to determine ad copy acceptance. 
The baseline mode scores relevance using only the query and ad copy. The context-aware mode enhances relevance scoring with RAG-retrieved product metadata. 
The generalization-aware mode incorporates both relevance and generalization, penalizing overly specific or vague outputs.
As shown in Table \ref{tab:evaluation_results}, the baseline mode exhibited the highest false negative rate (FNR, wasted rate) at 35.42\% and a false positive rate (FPR, concerned rate) of 1.53\%, indicating poor alignment with human evaluations. Adding context through the RAG-based mode significantly reduced FNR to 18.61\% while maintaining a similar FPR of 1.55\%. Incorporating generalization scores further reduced FPR to 0.97\% with stricter thresholds (4,3), though it increased FNR to 35.27\%, reflecting more rejected acceptable ad copies.
These results demonstrate that incorporating context and generalization scores improves alignment with human evaluations. This approach addresses safety concerns at scale while allowing threshold adjustments to balance both FPR and FNR based on evolving business priorities. 
Notably, using AutoEval-Main in place of human evaluation \textbf{reduces costs by 200× and processing time by 42×}, enabling scalable and efficient ad quality assessment.

\begin{table}[tbp]
\centering
\resizebox{0.49\textwidth}{!}{
\begin{tabular}{lccc}
\toprule
    \textbf{Method} & \textbf{FPR} & \textbf{FNR} & \textbf{Total} \\
    \textbf{(Thresholds)} &  \textbf{(Concerned)} & \textbf{(Wasted)} & \textbf{Disagree.} \\
    \midrule
    Keyword Only & 1.53\% & 35.42\% & 36.95\% \\
    Context Only  & 1.55\% & 18.61\% & 20.16\% \\
    Context (2) + General (3) & 1.53\% & \textbf{8.90\%} & \textbf{10.43\%} \\
    Context (3) + General (3) & 1.41\% & 15.25\% & 16.66\% \\
    Context (4) + General (2) & 1.03\% & 34.43\% & 35.46\% \\
    Context (4) + General (3) & \textbf{0.97\%} & 35.27\% & 36.24\% \\
    \bottomrule
\end{tabular}}
\caption{
Ablation study of false positive rate (FPR, concerned rate), false negative rate (FNR, wasted rate), and total misalignment rate between human and LLM evaluation across different methods and threshold settings.
``Context (X) + General (Y)'' denotes threshold settings, where X is the context relevance threshold and Y is the generalization threshold. 
}
\label{tab:evaluation_results}
\vspace{-10pt}
\end{table}

\subsection{Evaluation of AutoEval-Update} \label{sec: autoeval-update-eval}
Sampling strategies are crucial for ensuring that subsets selected for human review are both representative and meaningful. In this study, we evaluated AutoEval-Update using 1,000 keyword samples per strategy, comparing performance against a baseline represented by the original evaluation prompt. The baseline approach involves using the LLM-as-a-Judge without any refinement of the evaluation prompt, relying solely on initial scoring for ad copies.

The experimental design incorporated three sampling approaches, (1) random sampling; (2) uncertainty-based sampling; and (3) diversity based sampling.
The results, shown in Table~\ref{tab:auto_eval_update_results}, show that the baseline method performs the worst overall, with the highest FPR and FNR rates. All AutoEval-Update strategies improved upon the baseline to varying degrees. The optimal sampling strategy selection depends on specific objectives: uncertainty-based sampling proves most effective for reducing errors in complex cases, while diversity-based sampling offers superior coverage across various categorical and performance metrics.



To evaluate the performance of refined prompts and threshold settings, re-evaluations of prompt alignment are conducted using either of the two contamination-free datasets, meaning these datasets are not part of the
data used for report generation or prompt adjustment. (1) a validation set: used to test and iteratively refine adjustments. (2) a verified golden dataset: a reliable benchmark with no human error for assessing alignment and accuracy.
Before finalizing prompts, revalidation of the refined prompts is conducted, followed by a scientist check to confirm that the criteria and thresholds are robust. These steps form a systematic foundation for adapting AutoEval-Update to evolving data distributions and evaluation requirements.

The experimental results, obtained from simulations using previous human annotation data, are summarized in Table~\ref{tab:auto_eval_update_results} and Table~\ref{tab:autoeval-update-threshold}. These findings highlight three key insights: 
(1) \textbf{Role of Critic LLMs:} While critic LLMs effectively propose new evaluation criteria, human oversight remains crucial for defining and fine-tuning threshold settings (Table~\ref{tab:auto_eval_update_results}). 
(2) \textbf{Threshold Optimization:} The most reliable approach to setting thresholds involves prompt evaluations on a validation set, ensuring proper calibration before applying them at scale to enhance consistency and reliability.  
(3)  \textbf{Balancing Human-in-the-Loop and Automation:} Although human involvement is essential for maintaining evaluation quality, integrating automation with selective human review significantly reduces labor requirements (Table~\ref{tab:autoeval-update-threshold}).  
These findings demonstrate the importance of combining automated systems with minimal but strategic human oversight to achieve scalable, efficient, and high-quality evaluations.
A case study is provided in Appendix~\ref{appendix: case-study-autoeval-update}.

\begin{table}[tbp]
    \centering
    \resizebox{0.42\textwidth}{!}{
    \begin{tabular}{lcccc}
        \toprule
        \textbf{Method} & \textbf{FP} & \textbf{FN} & \textbf{Acc} & \textbf{F\textsubscript{beta}}\\
        \midrule
        Baseline (Original) & 20.69\% & 10.92\% & 68.39\% & 72.09\% \\
        Random  & 14.37\% & 15.52\% & 70.67\% & \textbf{73.07\%} \\
        Uncertainty & 17.82\% & 10.34\% & \textbf{72.69\%} & \textbf{73.43\%} \\
        Diversity & \textbf{13.22\%} & 22.41\% & 66.89\% & 70.16\% \\
        \bottomrule
    \end{tabular}}
    \caption{Results comparing original prompts and AutoEval-proposed prompts using various active sampling strategies.}
    \label{tab:auto_eval_update_results}
    \vspace{-10pt}
\end{table}

\begin{table}[tbp]
    \centering
    \resizebox{0.42\textwidth}{!}{
    \begin{tabular}{lcccc}
        \toprule
        \textbf{Method} & \textbf{FP} & \textbf{FN} & \textbf{Acc} & \textbf{F\textsubscript{beta}} \\
        \midrule
        AutoEval-Proposed & 26.44\% & 21.84\% & 51.72\% & 61.43\% \\
        ES - Max Recall & \textbf{14.37\%} & 15.52\% & 70.67\% & \textbf{73.07\%} \\
        ES - Balanced & 19.54\% & 8.62\% & 66.67\% & 70.16\% \\
        ES - Max Precision & 22.99\% & \textbf{5.75}\% & \textbf{71.26\%} & 71.97\% \\
        \bottomrule
    \end{tabular}}
    \caption{Ablation study comparing AutoEval-proposed thresholds with expert-suggested thresholds, using random sampling. ES stands for Expert-Suggested. }
    \label{tab:autoeval-update-threshold}
\end{table}

\subsection{Key Observations and Domain Insights }

\noindent\paragraph{Retriever Choice.} 
We initially employ a semantic-based retriever for RAG but find limitations in its effectiveness within the retail domain. In our 3,000-keyword experiment, human reviewers report a 10\% rejection rate for irrelevance and a 15\% overall rejection rate (detailed in Appendix~\ref{appendix: 3k-keyword-human-review}). A notable failure case is the "Stanley Cup problem," where the retriever confuses hockey games with trending water cups, leading to misleading ad copy.
To improve relevance, we adopt the search page products context retrieval strategy, which retrieves actual products from the e-commerce website’s search results. This approach significantly reduces the overall rejection rate to 2.79\% in human review, demonstrating improved contextual accuracy. We apply this retrieval method to both offline evaluations and the 10,000-keyword online experiment, ensuring more precise ad copy generation.

\noindent\paragraph{Domain-Specific Design.} General LLMs lack the nuanced understanding of ad network preferences and the specific criteria essential for optimizing ad performance. Our findings highlight the necessity of a domain-specific approach to ensure efficiency and effectiveness in this context. 
Insights from official recommendations provided by the advertising service reinforced the need for alignment with ad-specific requirements, prompting us to incorporate marketer-driven criteria into our evaluation framework.
By integrating marketer insights (detailed in Appendix~\ref{appendix: marketer_insight}), we ensure that our system is tailored to the unique demands of digital advertising, ultimately improving ad quality and performance in real-world applications.

\noindent\paragraph{Automated Evaluation.} 
Our experience with LLM-as-a-Judge highlights the importance of starting with \textit{high-quality, large-scale human annotations} to establish a reliable benchmark for automated evaluation. This foundation enables robust simulation and validation, ensuring that the system aligns closely with human judgment.
One notable finding is that no safety or toxicity issues are detected in the generated content, reinforcing the system’s trustworthiness. 
Instead, the key challenge is defining evaluation criteria that effectively filter out low-quality ad copies while minimizing unnecessary rejections, emphasizing the need for precise, well-calibrated standards to optimize ad quality and relevance.

\noindent\paragraph{Self-Refinement of Evaluation Prompt for Criteria Drifts.}
As data distributions and evaluation criteria evolve, it is imperative to develop adaptive systems that can dynamically adjust to these shifts. Our human-LLM collaborative framework enables continuous refinement by incorporating newly emerging assertions and criteria, ensuring that evaluations remain aligned with evolving marketing principles. Moreover, the efficiency and reliability of this approach hinge on an optimized balance between human oversight and automation. While human-in-the-loop validation remains essential for maintaining accuracy, our findings demonstrate that automation and selective sampling can significantly reduce manual effort. With this cost-efficient LLM-human collaboration, we establish a scalable and adaptive evaluation system that sustains high-quality assessments with minimal human intervention.


\section{Conclusion And Future Work} 
In this work, we introduce a scalable framework for retrieval-augmented ad copy generation and evaluation on an e-commerce website, addressing the challenges posed by dynamic product catalogs, diverse customer intents, and policy compliance in large-scale marketing. By grounding content generation in real-time product data, employing channel-specific constraints, and combining rule-based checks with LLM evaluations, the proposed approach achieves enhanced precision, reduced error rates, and substantial cost savings in manual annotation processes. The experimental results validate the system's effectiveness, demonstrating significant improvements in engagement metrics such as clicks and sales.
For future work, we aim to integrate our system into an LLM agent to facilitate expert interactions for system refinement, explore model distillation to enhance efficiency and continue training to further improve performance.

\bibliographystyle{plainnat}
\bibliography{references}

\newpage
\newpage
\appendix

\section{ Prompt Examples \label{sec:Appendix:prompts-examples}}
\subsection{General Generation} \label{appendix: prompt-gen}
The marketing content generation prompt example used for our partner major search engine advertising service use case is as follows:
\begin{Verbatim}[frame=single, breaklines=true]
Human: 

Context:
{context}

Let's think step by step.

Follow the below instructions when generating answers:
{detailed_ads_content_requirements_and_instructions}

Here are some examples. Each example has a question or the keyword and corresponding headlines and descriptions.
{examples}

Question: {question}

Assistant:
\end{Verbatim}

\subsection{Call-to-Action Generation.}
\begin{Verbatim}[frame=single, breaklines=true]
Human: 
    
Lets think step by step.
 
Follow the below instructions when generating answers:
{detailed ads content requirements and instructions for call to action, e.g., 
Add "at [e-commerce website]" at the end of headlines when question length is short;
Each headline must start with call-to-action verbs. Choose approporiate verb based on the question.}

Here are some examples. Each example has a question or the keyword and corresponding headlines.
{examples}

Question: {question} 

Assistant:

\end{Verbatim}

\subsection{Task Chaining Rewrite} \label{appendix: prompt-rewrite}
The task chaining rewrite prompt for the Paid Search search engine advertising service use case is as follows:
\begin{Verbatim}[frame=single, breaklines=true]
Human:

Use headlines and descriptions in <rewrite></rewrite> XML tags below to create shorter versions. 

Previous Output:
<rewrite>
{previous_output}
</rewrite>

Let's think step by step.

The headlines and the descriptions are previously generated by an LLM but the LLM did not follow all our requirements in generating headlines and descriptions.
Your task is to fix the issues based on the following instructions and rewrite new headlines and descriptions.


Instructions:
{detailed ads content requirements and instructions}

Question: {question} 

Assistant:
\end{Verbatim}

\subsection{AutoEval-Main} \label{appendix: evaluation_prompt}
The automated evaluation prompt for the [our partner search engine advertising service] use case is presented as follows:
\begin{Verbatim}[frame=single, breaklines=true]
    Human:
        Use ad copies generated by another LLM for a search query to evaluate their quality and provide scores.
        The content is ad copies for a search query.
        The search query is provided in <keyword></keyword> XML tags.
        The search query has associated context, provided below in <context></context> XML tags.
        The content is provided below in <content></content> XML tags.
        Each ad copy is inside <ad_copy></ad_copy> XML tags inside <content></content> XML tags.

        For each ad copy, you need to follow these steps:
        Step 1: Analyze the user query: <keyword>{keyword}</keyword>. Consider its context: <context>{context}</context>.
        Step 2: Analyze the response: {content}
        Step 3: Evaluate the generated response based on the following criteria and provide a score from 0 to 5 along with a brief justification for the criterion:
        {designed criteria}
        
It should avoid being:
{designed rejection reasons}

Here are some examples.
{examples}
            
Assistant:
"""
\end{Verbatim}

\begin{figure*}[bp]
    \centering
    \includegraphics[width=\textwidth]{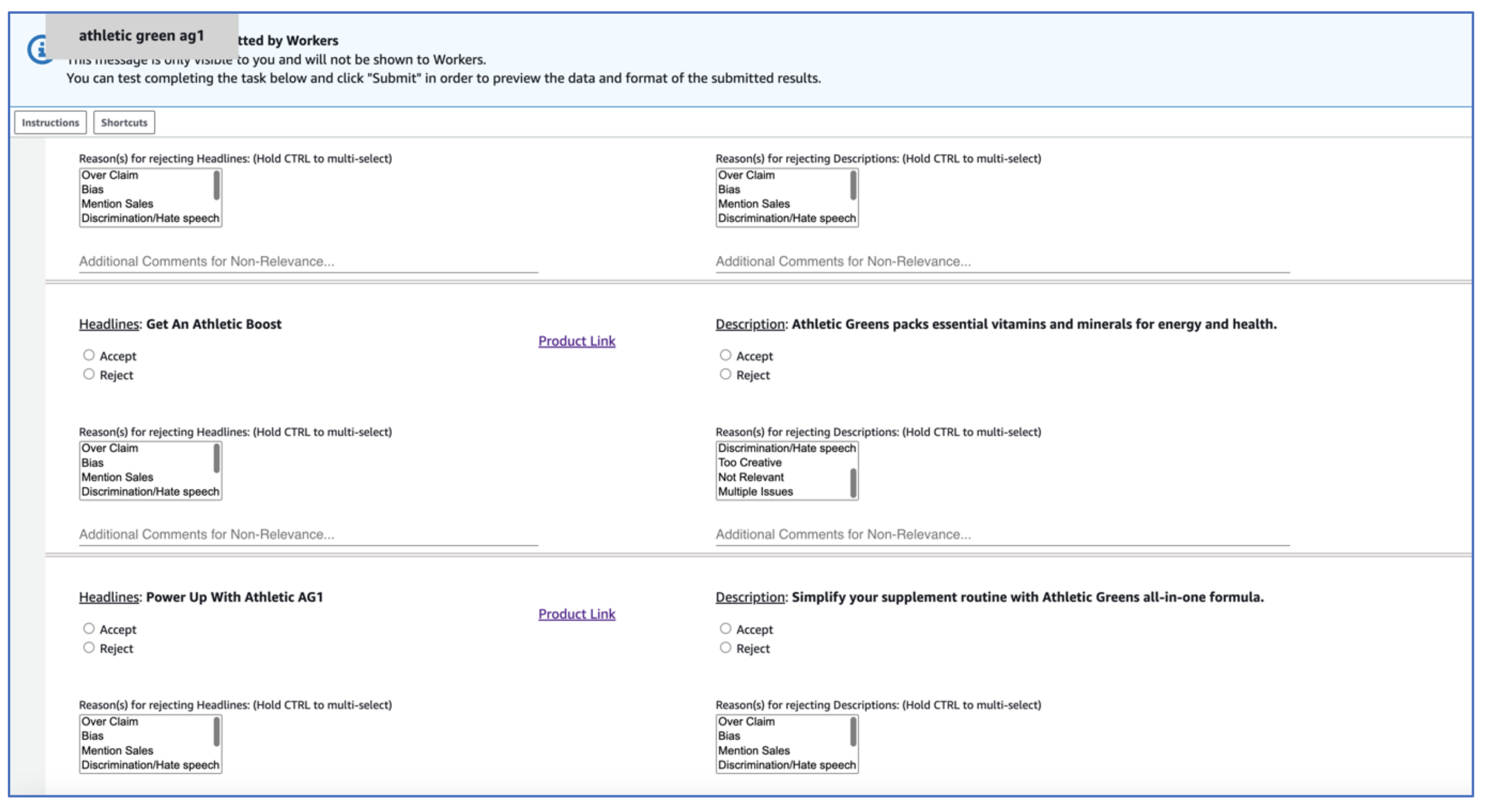}
    \caption{AWS GroundTruth UI for human annotation of MarketingFM generated marketing content.}
    \label{fig:validation_UI.png}
\end{figure*}


\section{Human Evaluation}
\subsection{Marketer Insight for Ads Quality} \label{appendix: marketer_insight}
To ensure high-quality ad copy, we collaborate with marketers and identify three key aspects: \textbf{relevance}, \textbf{call to action (CTA)}, and \textbf{dynamic keyword insertion (DKI)}. These factors play a crucial role in optimizing engagement and alignment with marketing goals.  

\paragraph{Relevance}  
To assess relevance, we provided LLM-generated headlines to , which scored their alignment with the search keyword on a scale from 0 to 5, where 0 indicates no relevance and 5 represents high relevance. Headlines scoring below 4 were rejected and replaced to maintain ad quality.  

\paragraph{Call to Action (CTA).}  
Effective ad copy encourages user engagement through action-oriented language. We checked for the presence of CTA verbs such as \textit{“Buy Now”} or \textit{“Shop at [the e-Commerce website]”}. Headlines lacking strong CTA phrases were removed and regenerated to improve conversion potential.  

\paragraph{Dynamic Keyword Insertion (DKI).}  
DKI enhances ad relevance by incorporating the search keyword directly into the ad text. While we did not enforce strict rules regarding DKI usage or headline length—given variations in keyword length—we analyzed the generated output to understand their distribution and impact.  

These insights guide our approach to refining ad content, ensuring that LLM-generated ads meet marketing standards and drive engagement effectively.

\subsection{AWS GroundTruth UI Developed for Validation\label{sec:Appendix:validation-ui}}  

As shown in Figure~\ref{fig:validation_UI.png}, we developed a custom validation interface using AWS GroundTruth to streamline the human evaluation process. This UI enables annotators to efficiently review and assess LLM-generated ad copies based on predefined quality guidelines.  

The interface presents each ad copy alongside its associated keyword and product information, allowing annotators to make informed decisions. Annotators can mark an ad copy as \textit{accepted} or \textit{rejected} and, in the case of rejection, select from predefined rejection reasons (e.g., \textit{not relevant}, \textit{lacking CTA}, \textit{misleading claim}).  

To ensure consistency, annotators receive training based on detailed guidelines, and their decisions are periodically audited. The validation tool also logs annotation metadata, enabling further analysis of inter-annotator agreement and evaluation trends.

\subsection{Human Evaluation Rejection Criteria} \label{appendix: human_rejection_criteria}
To ensure high-quality ad copy, human evaluators follow a structured rejection framework based on predefined criteria. Table~\ref{tab: human_rejection_reasons} outlines the primary reasons for rejection, each aimed at maintaining relevance, clarity, and compliance with marketing standards.

\begin{table*}[h!]
\centering
\renewcommand{\arraystretch}{1.05} 
\begin{tabular}{p{2cm}p{12cm}}
\toprule
\textbf{Reason} & \textbf{Definition} \\
\toprule
Not Relevant & The ad copy is unrelated or only loosely connected to the keyword, promoting a different product or feature. \\
\midrule
Too Specific & The ad copy is too narrowly focused on a particular feature or product, limiting its relevance to the broader keyword intent. \\
\midrule
Over Claim & The ad copy exaggerates the product's features or benefits beyond what is reasonable or substantiated. \\
\midrule
Mention Sales & Ad copies must avoid indication/references to sales, discounts, or promotional terms, focusing instead on product features and benefits. \\
\midrule
Bias & The ad copy contains assumptions or stereotypes based on gender, age, or other demographics, potentially excluding certain audiences. \\
\midrule
Too Creative & The ad copy is overly abstract or imaginative, straying from clear and direct messaging, potentially confusing or misaligning with the keyword intent. \\
\midrule
Others & The ad copy is vague, unclear, or incomplete, failing to align with any predefined rejection categories. \\
\midrule
Hate Speech & The ad copy contains offensive or harmful language targeting specific groups. \\
\bottomrule
\end{tabular}
\caption{Reasons for rejection with definitions.}
\label{tab: human_rejection_reasons}
\end{table*}

\subsection{Human Evaluation Results for First-Round 3,000-Keyword Test}\label{appendix: 3k-keyword-human-review}
\begin{figure*}[h!]
    \centering
    \includegraphics[width=0.6\textwidth]{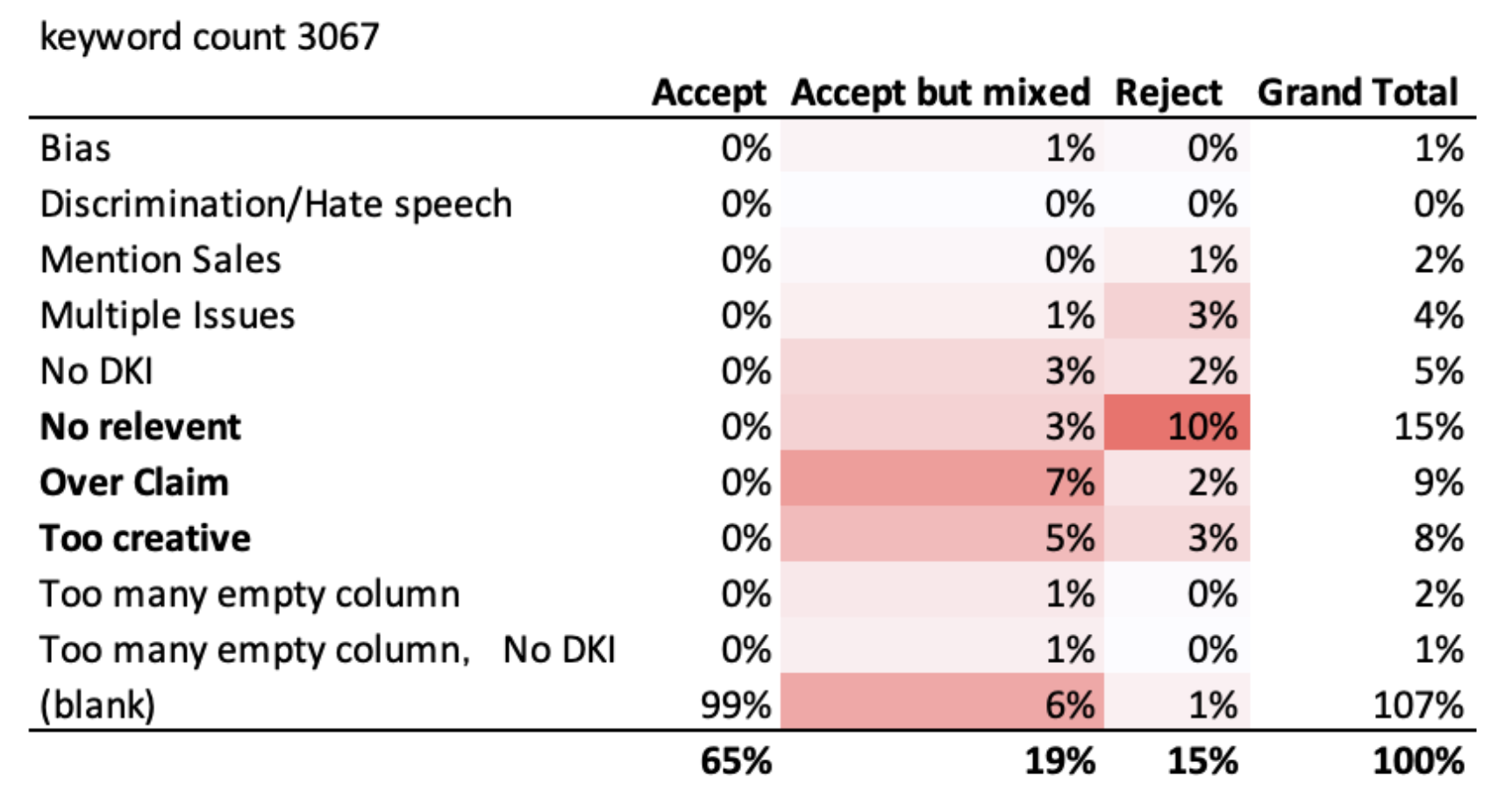}
    \caption{Human evaluation results for the first-round 3,000-keyword test using semantic embedding-based RAG generation.}
    \label{fig: human-eval-reasons-3k}
\end{figure*}

Figure~\ref{fig: human-eval-reasons-3k} presents the human review results for the 3,000-keyword initial test using the semantic-based retriever. The overall rejection rate is 15\%, with "Not Relevant" (10\%) being the most frequent issue, followed by "Over Claim" (7\%) and "Too Creative" (5\%). These findings indicate that the semantic-based retrieval approach often fails to retrieve contextually relevant ads, leading to a high rate of irrelevant or misleading ad copies.

\section{AutoEval-Update Framework} \label{appendix: autoeval-update-framework}
\begin{figure*}[h!]
    \centering
    \includegraphics[width=0.8\textwidth]{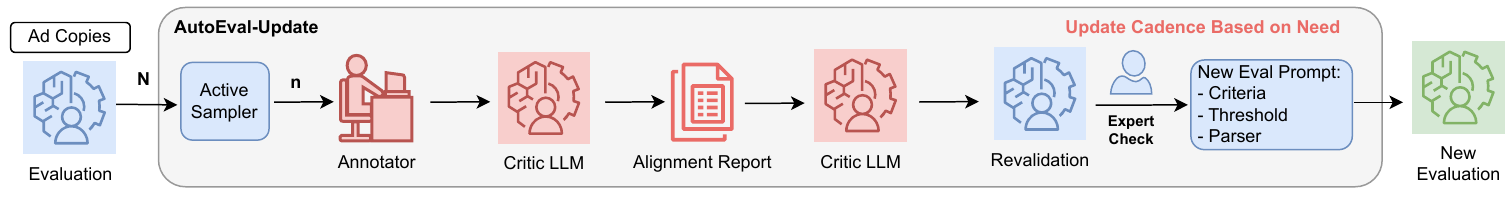}
    \caption{An illustration of the AutoEval-Update pipeline for self-refinement of evaluation prompt.
    }
    \label{fig:autoeval-update-framework}
\end{figure*}

Figure~\ref{fig:autoeval-update-framework} illustrates the AutoEval-Update framework, which iteratively refines evaluation prompts to maintain alignment with evolving data and criteria. The framework integrates active sampling, human review, and LLM-based evaluation refinement to systematically and reliably improve assessment accuracy while minimizing necessary human and expert labor.

\section{Case Study: Evaluation Prompt Refinement with AutoEval-Update} \label{appendix: case-study-autoeval-update}

\begin{figure*}[h!]
    \centering
    \includegraphics[width=0.8\textwidth]{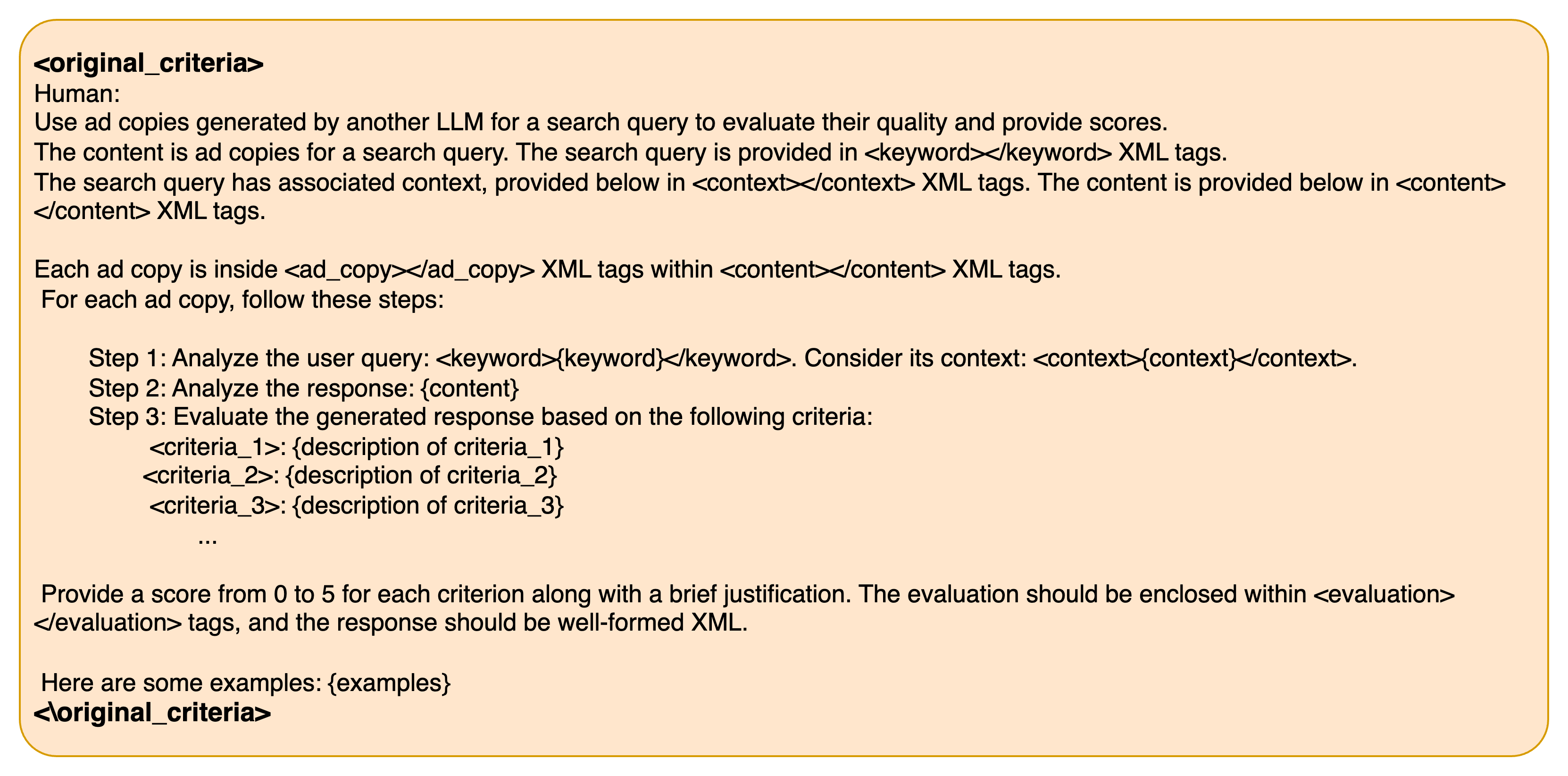}
    \caption{Initial evaluation prompt before refinement by AutoEval-Update.}
    \label{fig:autoeval-update-old}
\end{figure*}

\begin{figure*}[h!]
    \centering
    \includegraphics[width=0.8\textwidth]{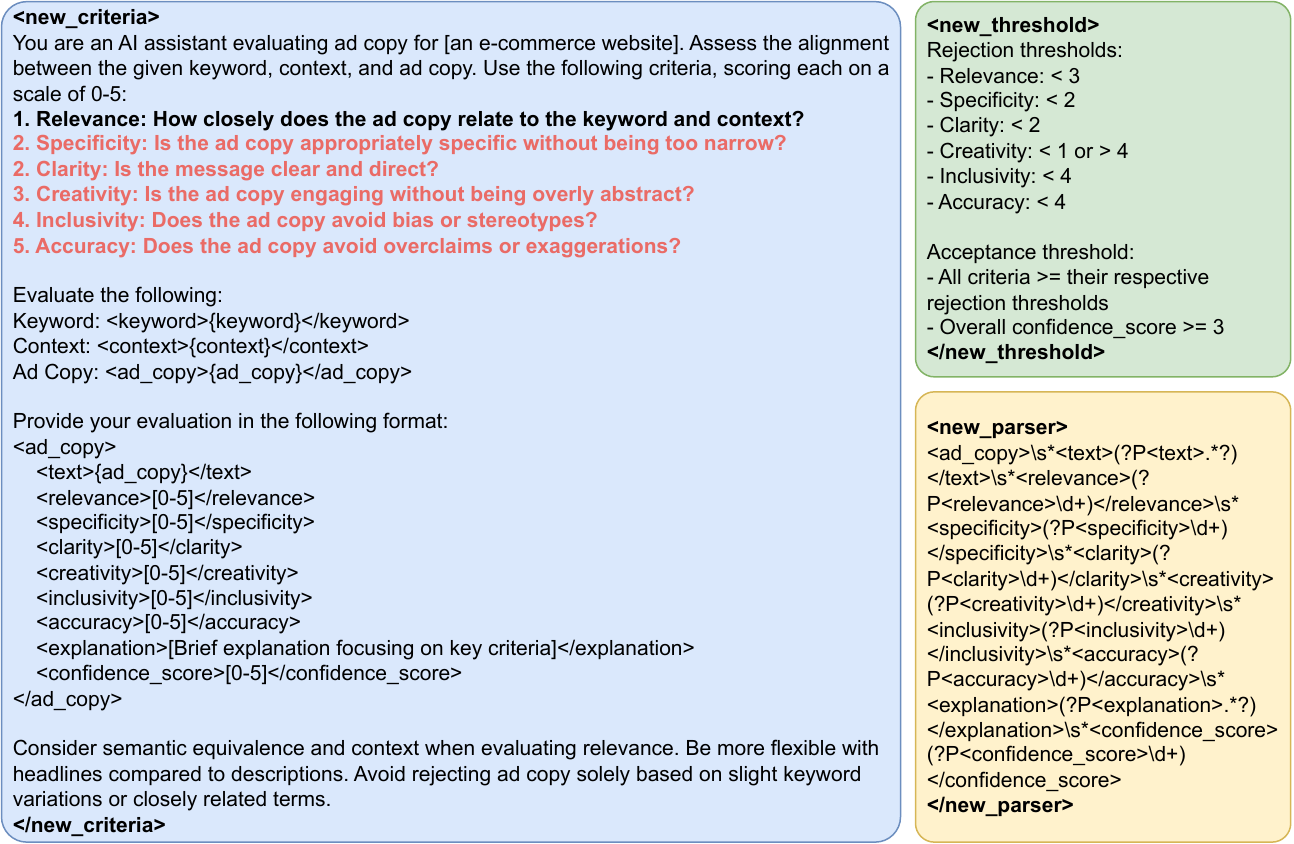}
    \caption{Refined evaluation prompt generated through AutoEval-Update.}
    \label{fig:autoeval-update-new}
\end{figure*}

In this section, we present a case study on evaluation prompt refinement using diversity sampling with AutoEval-Update, simulated with our large-scale human-annotated data.  
Figure~\ref{fig:autoeval-update-old} shows the \textit{original} evaluation criteria for ad copies, which included DKI, CTA, relevance, and mention sale, along with their associated limitations.  
Figure~\ref{fig:autoeval-update-new} presents the \textit{refined} criteria, incorporating additional dimensions, proposed acceptance thresholds, and a structured parser to extract scores from the LLM-as-a-Judge output.  
This shows that AutoEval-Update can provide useful and insightful evaluation criteria. 

\begin{figure*}[tbp] 
    \centering
    \includegraphics[width=0.8\textwidth]{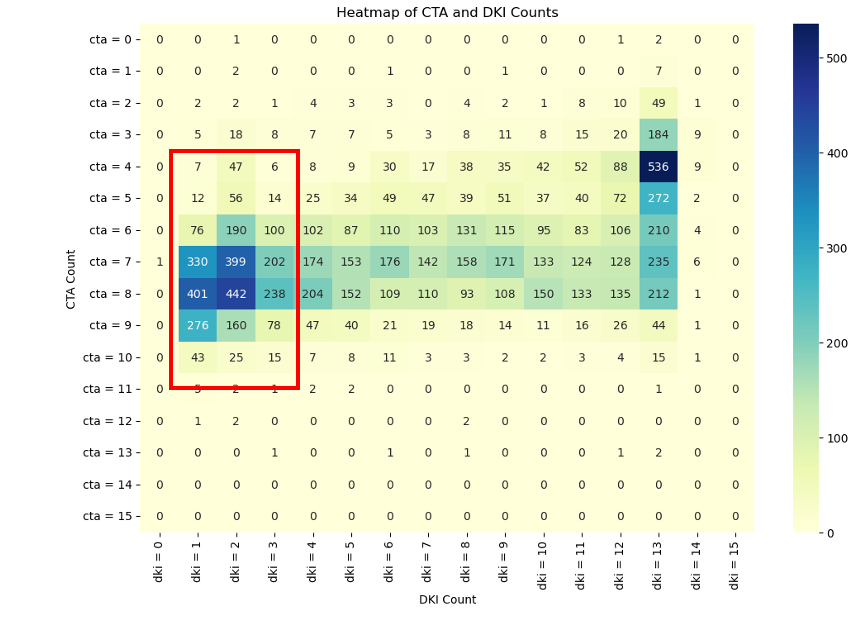}
    \caption{A heatmap between CTA and DKI counts.
    The number of keywords in the sample of 10,000 keywords with the number of DKI and CTA headlines. The numbers in the boxes represent the number of keywords. The x-axis is the number of headlines that meet the DKI requirement, and the y-axis represents the number of headlines that meet the CTA requirements.}
    \label{fig:CTA_DKI_heatmap}
\end{figure*}

\section{More on DKI and CTA Metrics}
\label{appendix:cta_dki}
While we cannot enforce DKI due to length limitations, we looked at the number of DKI and CTA headlines and how many keywords exist for each combination, as shown in Figure~\ref{fig:CTA_DKI_heatmap}. On the left side of the heat map with less than 3 DKI headlines (red box), the average length of keywords is 24 characters, which means it is harder to increase the number of DKI when the keyword is too long.

\end{document}